# A novel hybrid methodology of measuring sentence similarity

Yongmin Yoo°, Tak-Sung Heo°, Yeongjoon Park° and Kyungsun Kim*

°Equal contribution
NHN Diquest

yooyongmin91@gmail.com
gjxkrtjd221@gmail.com
yeongjoon1227@gmail.com
kksun@diquest.com

## Abstract

The problem of measuring sentence similarity is an essential issue in the natural language processing area. It is necessary to measure the similarity between sentences accurately. There are many approaches to measuring sentence similarity. Deep learning methodology shows a state-of-the-art performance in many natural language processing fields and is used a lot in sentence similarity measurement methods. However, in the natural language processing field, considering the structure of the sentence or the word structure that makes up the sentence is also important. In this study, we propose a methodology combined with both deep learning methodology and a method considering lexical relationships. Our evaluation metric is the Pearson correlation coefficient and Spearman correlation coefficient. As a result, the proposed method outperforms the current approaches on a KorSTS standard benchmark Korean dataset. Moreover, it performs a maximum of 65% increase than only using deep learning methodology. Experiments show that our proposed method generally results in better performance than those with only a deep learning model.

## 1. Introduction

As natural language data such as social networks and news articles are pouring out, natural language processing has received tremendous attention [1]. Measuring the similarity between natural language sentences is important even within natural language processing [2]. For example, many approaches such as chat-bot system, plagiarism checking system, automatically classification system depend on sentence similarity. Accurately measuring the similarity between two sentences is a crucial task.

Researches measuring sentence similarity have been conducted from various perspectives [2-6]. Including deep learning approaches and sentence structure approaches, there are many ways to measure sentence similarity.

In a study on sentence similarity measurement using deep learning, Mueller et al. [6] proposed a method by extracting features containing the entire information of a sentence via long short-term memory (LSTM). Heo et al. [7] proposed a method to measure sentence similarity using a combination of global features: the entire information of sentences extracted via bidirectional LSTM (Bi-LSTM) and local features: the detailed information of sentences extracted through capsule networks.

In the natural language processing field, it is necessary to focus not only on the model using deep learning but also on the structure of the sentence and the lexical relationship of the sentence. Miller et al. [4] proposed a method measuring similarity between two sentences based on the relationship between vocabulary within sentences, using WordNet made from a knowledge base using vocabulary. In contrast, Wang et al. [5] proposed a method of decomposing and reorganizing vocabulary.

In this paper, our research aims to improve the performance of methods for measuring similarity between the Korean language sentences by combining a deep learning methodology and a methodology that considers lexical relationships. We use several deep learning methodologies, such as convolutional neural networks (CNN), recurrent neural networks (RNN), and bidirectional encoder representations from transformers (BERT) to measure sentence similarity. Also, we apply cosine similarity to embedding vectors obtained from the language representation model. Finally, we calculate the final sentence similarity by combining the sentence similarity value calculated by the deep learning model and the value obtained from cosine similarity. Experiments show that our proposed method performed better compared to those with only a deep learning model.

This paper is structured as follows. Section2 mentions related work. Section3 explains the proposed approach and details of its main components. Section4 describes the experiment, Section5 mentions the conclusion.

## 2. Related Work

Many approaches have been proposed to address the problem of measuring the similarity between sentences [9]. Research measuring the similarity between two sentences has been conducted for a long time from various perspectives. There are many approaches to calculate the similarity between two sentences, such as using sentence structure, considering a lexical relationship, and using deep learning.

The method of measuring the similarity between two sentences using the structure of sentences is a widely used method from the early days of natural language processing to the present era of deep learning. Since many researchers have studied it for a long time, many ideas have been proposed to measure the similarity between two sentences using sentence structure. Lee et al. [10] proposed a similarity measure method of two sentences based on sentence structure grammar. As Lee et al. [11] proposed a similarity measure method of two sentences based on the part of speech tags. Ferreira et al. [2] proposed a similarity measurement method of two sentences based on the word order and sentence structure. Li et al. [3] measured the similarity of two sentences by identifying statistics on sentence structure.

A method of measuring similarity between two sentences considering lexical relationships is also one of the sentence similarity measures. Miller et al. [4] proposed a method for measuring similarity between two sentences using WordNet made from a lexical-based knowledge base. Wang et al. [5] are presented with a method to calculate the similarity between two sentences using an approach that uses the word to calculate similarity using repetitive and different parts of a sentence. Abdalgader et al. [12] proposed a method to measure sentence similarity using word detection ambiguity and synonym extension.

Deep learning has recently developed significantly since hardware development and the opening of the Big Data era [13]. Sentence similarity studies using deep learning have shown good performance using various neural networks such as LSTM, gated recurrent units (GRU), CNN, and BERT [6-8, 14, 15].

Mueller et al. [6] used LSTM, which has a good performance for sequential data processing. They evaluated sentence similarity by applying the last hidden states extracted via LSTM to Manhattan distance. Pontes et al. [14] combined CNN and LSTM. They extract combined information from adjacent words through CNN and applied last hidden states extracted via LSTM to Manhattan distances to assess sentence similarity. Li et al. [15] used Group CNN (G-CNN), which extracts representative local features and bidirectional GRU (Bi-GRU), which has good performance in sequential data processing and applied last hidden states extracted via Bi-GRU to Manhattan distances. Heo et al. [7] sequentially used Bi-LSTM, self-attention reflecting contextual information, capsule networks with CNN structure. They then combined the last hidden states extracted via Bi-LSTM and local features extracted via capsule networks. Devlin et al. [8] evaluated the similarity of the two sentences using BERT, a language representation model that shows excellent performance in various natural language processing fields.

Unlike previous approaches, we propose a novel method that combines deep learning and the method considering lexical relationships to measure similarity between two sentences.

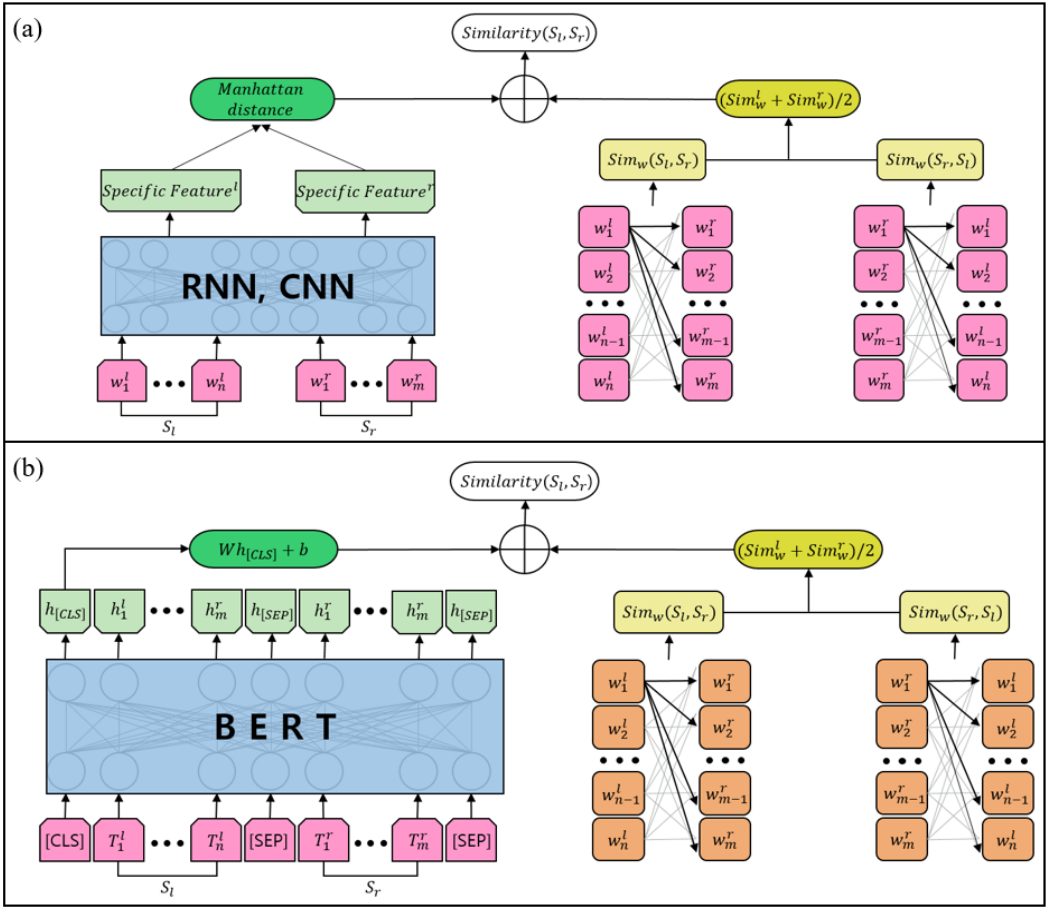

Fig. 1. Proposed system architecture: (a). Sentence similarity hybrid model using Word Embedding, (b). Sentence similarity hybrid model using BERT

### 3. Sentence Similarity

Measuring accurately in measuring the similarity of sentence to sentence is an important task [2]. To measuring similarity between sentences, we combine deep learning methodology and a method that considers lexical relationships.

### 3.1 Similarity based on Deep Learning Model

The models used in this study are LSTM and GRU, CNN, G-CNN, capsule networks, BERT. LSTM and GRU are a family of RNN, and G-CNN and capsule networks are neural networks using CNN. CNN and G-CNN are used as input values to the RNNs model. The sentence similarity using RNNs or capsule networks is calculated by applying the Manhattan distance, such as Fig. 1 (a). And, the sentence similarity using BERT is calculated through a special token, such as Fig. 1 (b).

### 3.1.1 Sentence similarity using Word Embedding

The RNN is a neural network that shows good performance when processing sequential data [16]. When calculating the representation of each time step in text processing, it is determined through learning how much context information up to that point is reflected. However, in the case of RNN, if the sequence length is increased, gradient vanishing or gradient exploding problems may occur [17]. To solve this, LSTM and GRU are devised [17, 18]. When calculating the similarity of two sentences, each sentence is input into the family model of RNNs to obtain the last hidden states ($h^l_L, h^r_L$) of each sentence [6-7, 14-15]. The $h_L$ includes the entire sentence information. Manhattan distance used in the family model of RNNs is as follows.

$$Sim_D = \exp\left(-\left\|h^l_L - h^r_L\right\|_1\right) \quad (1)$$

The CNN shows good performance in image and text processing [7]. In the text processing, CNN extracts local features, which are combined information, by grouping words appearing in sentences by kernel size ($k$), and increases the amount of learning by using the number of filters ($N$). The equation of CNN is as follows.

$$Conv_i = f(W_k * w_{i-\frac{k-1}{2}:i-\frac{k+1}{2}} + b_k) \quad (2)$$

In equation (2), $i$ refers to word index, $f(\cdot)$ refers to the activation function. In addition, $W_k$ refers to a learning weight of a CNN having a size of $\mathbb{R}^{k \times E}$, $w_i$ refers to a word embedding vector having a size of $\mathbb{R}^E$, and $b_k$ refers to a bias vector. Through equation (2), the feature map, which is having a size of $\mathbb{R}^{L-k+1}$, is generated according to $L$, which refers to the number of words. Finally, the feature maps having a size of $\mathbb{R}^{(L-k+1) \times N}$ are generated by the number of filters.

The G-CNN uses three CNNs with different kernel sizes in parallel and obtains representative semantic information [15]. G-CNN integrates feature maps extracted from three CNNs into a feature of one and then creates the most representative feature map by applying max pooling. The equation of G-CNN is as follows.

$$G\text{-}CNN_i = \max(CNN_i^k, CNN_i^{k+2}, CNN_i^{k+4}) \quad (3)$$

The capsule networks use two $Conv1$ and $Conv2$ and are used to extract subdivided information in the field of sentence similarity [7]. $Conv1$ has a typical CNN form, and $Conv2$ receives a feature map of $Conv2$ as the input value. After that, to extract subdivided information, a kernel size corresponding to the overall size of the input value is used, and then a feature map having the size of $\mathbb{R}^{C \times \frac{N}{C}}$ is generated by dividing the feature map into $C$-dimensions. Manhattan distance used in the capsule networks is as follows.

$$Sim_D = \frac{1}{C}\sum_i^C exp(-\left\| \begin{matrix} Conv2^l_i \\ -Conv2^r_i \end{matrix} \right\|_1) \quad (4)$$

### 3.1.2 Sentence similarity using BERT Embedding

BERT is a language representation model made by stacking several transformer encoder blocks [8]. The learning process of BERT is divided into a pre-training process and a fine-tuning process. In the pre-training process, after randomly masking word tokens in a sentence of a large corpus, the BERT model is learned by predicting the masked word token. Fine-tuning is a process of learning a pre-trained BERT model with labeled data once more. We train BERT on the similarity task between sentences in the fine-tuning process.

BERT is divided into BERT-base and BERT-large models according to the size of the model, and in this work, we use the BERT-base model. BERT-base model consists of 12 layers of transformer encoder block. There are several special tokens in the BERT model. The [CLS] token is placed at the beginning as a token indicating the beginning of the input. The [SEP] token is a token that distinguishes between sentences. In the case of the sentence similarity task, the input is entered in the BERT model in the form of "[CLS] Sentence1 [SEP] Sentence2 [SEP]". The similarity between sentences is calculated by inputting the output vector of [CLS] token extracted via the BERT model into the dense layer, as shown in equation (5).

$$Sim_D = \sigma(Wh_{[CLS]} + b) \quad (5)$$

The $h_{[CLS]}$ is the vector representation of the [CLS] token extracted through the BERT model and has a matrix of $\mathbb{R}^{728}$. $W$ refers to weights that can be learned, with a matrix of $\mathbb{R}^{1 \times 728}$, and $b$ refers to bias vector.

### 3.2 Similarity based on lexical relationship

To measuring structure similarity, the proposed approach calculates the word-to-word similarity between sentences. Using language representation model in measuring similarity between words improves similarity measuring between sentences [19].

In the word embedding model, to consider the lexical relationship included semantic information, we use the embedding vector of words ($w_n$). Word embedding used in this study uses Word2Vec, which learns by minimizing the dot product values of target word vector and neighbor word vectors surrounding the target word vector [20, 21].

In the BERT, to consider the lexical relationship included contextual information, we use the hidden states of the 12 transformer blocks, which are BERT-base model components. To calculate the word-to-word similarity, we exclude hidden states of [CLS] token and [SEP] token. The embedding token of the word ($w_n$) is calculated using the following equation.

$$w_n = (\sum_{i=1}^{T} h_{i,n})/L \qquad (6)$$

In equation (6), $n$ refers to the index of the word, $T$ refers to the number of transformer blocks.

The word-based similarity is calculated using the following equation.

$$sim_w(S_l, S_r) = \left(\sum_{i=1}^{L} sim(w_i^l, S_r)\right)/n \qquad (7)$$

In equation (7), $S_l$ and $S_r$ refer to sentence 1 and sentence 2, respectively. $L$ is the number of words in $S_l$ and the $sim(w_i, S_r)$ is the similarity of word $w_i$ in $S_l$ and $S_r$. The similarity between word $w$ and $S$ is measured by selecting the max similarity between $w$ and every word in the $S$ according to the following equation (8).

$$sim(w, S) = \max(sim(w, w_j)) \qquad (8)$$

In equation (8), $sim(w, w_j)$ is the similarity between the word $w$ and word $w_j$. $w_j$ refers to the $j$th word that appears in another sentence $S$. The similarity between words is measured using cosine similarity between word vectors.

Our method goes through the process of calculating the similarity of words belonging to $S_r$ based on words belonging to $S_l$, as in the above formula, when comparing two sentences $S_l$ and $S_r$, and vice versa. In other words, we calculate the word in $S_l$ based on the word in $S_r$ and then calculate the arithmetic mean of the method, such as equation (9), from two methods.

$$Sim_w = \frac{sim_w(S_l, S_r) + sim_w(S_r, S_l)}{2} \qquad (9)$$

### 3.3 Novel hybrid sentence similarity

In this study, we combine deep learning methodology and a method that considers lexical relationships using the equation below.

$$Similarity = \alpha Sim_D + (1-\alpha) Sim_w \qquad (10)$$

In equation (10), α is a weight that can adjust which information to focus on among deep learning methodology and a method that considers lexical relationships and is determined experimentally. Through Equation (10), the value of sentence similarity has a range of 0-1.

## 4. Experiment

BERT used in this study is KoBERT[1], which trained Korean texts. And, Word2Vec is trained using the Korean raw corpus[2] applied Kkma[3], a Korean morpheme analyzer. Some studies have shown that splitting words into morphemes in Korean tends to perform well [22]. Word vector extracted through Word2Vec has an embedding size of $\mathbb{R}^{768}$, as same as BERT embedding size.

### 4.1 Datasets

In this study, KorSTS [23], consisting of 8,628 sentence pairs, is used as the experiment data. Training sets, development sets, and test sets consist of 5,749, 1,500, and 1,379 sentence pairs. The similarity score range of two sentences is composed from 0 to 5 points, as shown in Table 1.

Table 1: KorSTS dataset example

| Sentence 1 | Sentence 2 | Score |
|---|---|---|
| 한 여성이 브로콜리를 자르고 있다. "A woman is cutting broccoli." | 한 여성이 칼로 브로콜리를 자르고 있다. "A woman is cutting broccoli with a knife." | 4.25 |
| 남자가 카누로 노를 저고 있다. "The man is rowing a canoe." | 남자가 하프를 연주하고 있다. "The man is playing the harp." | 0.667 |

---

[1] https://github.com/kiyoungkim1/LMkor
[2] http://nlp.kookmin.ac.kr/kcc/
[3] http://kkma.snu.ac.kr/

Table 2: Performance comparison

| Model | Deep Learning | | Hybrid | |
|---|---|---|---|---|
| | Pearson | Spearman | Pearson | Spearman |
| LSTM [6] | 0.362511 | 0.344804 | **0.557284** | **0.550898** |
| CNN + LSTM [14] | 0.35528 | 0.334058 | **0.556268** | **0.55126** |
| G-CNN [15] | 0.604134 | 0.577296 | **0.65184** | **0.640162** |
| Capsule [7] | 0.620881 | 0.599617 | **0.661728** | **0.65319** |
| BERT [8] | 0.838989 | 0.830135 | **0.842797** | **0.834181** |

In this study, the similarity score is normalized from 0 to 1 using the minimum-maximum scaling.

**4.2 Result**

We compare the five models that perform high performance and the proposed model. We use the Pearson correlation coefficient and the Spearman correlation coefficient as the evaluation metrics.

The performance shown in Table 2 is the average value of the results of the 5 experiments each. As shown in Table 2, we can see that the Pearson and Spearman correlation coefficients are both higher in consideration of both deep learning and lexical relationships than those using only deep learning.

Our method resulted in significant performance improvement, though minor calculation cost increase compared to those models only using deep learning. Given that we achieved performance improvements in all five models, we show that our method generally increases the ability of the model to understand the semantic similarity of sentences. Especially on [14]'s model, our approach resulted in about 65% performance improvement.

**5. Conclusion**

This study measures the similarity between the Korean language sentences by combining a deep learning methodology and a method that considers lexical relationships. In a deep learning methodology, we use 5 neural networks related CNN, RNN, BERT. Also, in a method that considers lexical relationships, we use a cosine similarity in embedding vectors extracted through the word representation model. Finally, we calculate the final sentence similarity by combining the output values of the two methods. As a result, our method combining two methods shows good performance compare to using only a deep learning model.

The method considering lexical relationships used in this study is one of several linguistic methods for measuring sentence similarity. As it can see from the experimental results, it can achieve good performance by combining the deep learning method and the linguistic method. Therefore, in future studies, we will further improve the performance of similarity between the sentences by using various linguistic methods such as information on the order of words and part of speech.

**Reference**

[1] Hirschberg, Julia, and Christopher D. Manning. "Advances in natural language processing." Science 349.6245 (2015): 261-266.

[2] Ferreira, Rafael, et al. "Assessing sentence similarity through lexical, syntactic and semantic analysis." *Computer Speech & Language* 39 (2016): 1-28.

[3] Li, Yuhua, et al. "Sentence similarity based on semantic nets and corpus statistics." *IEEE transactions on knowledge and data engineering* 18.8 (2006): 1138-1150.

[4] Miller, George A. "WordNet: a lexical database for English." *Communications of the ACM* 38.11 (1995): 39-41.

[5] Wang, Zhiguo, Haitao Mi, and Abraham Ittycheriah. "Sentence similarity learning by lexical decomposition and composition." *arXiv preprint arXiv:1602.07019* (2016).

[6] Mueller, Jonas, and Aditya Thyagarajan. "Siamese recurrent architectures for learning sentence similarity." Proceedings of the AAAI Conference on Artificial Intelligence. Vol. 30. No. 1. 2016..

[7] Heo, Tak-Sung, et al. "Global and Local Information Adjustment for Semantic Similarity Evaluation." *Applied Sciences* 11.5 (2021): 2161.

[8] Devlin, Jacob, et al. "Bert: Pre-training of deep bidirectional transformers for language understanding." *arXiv preprint arXiv:1810.04805* (2018).

[9] Farouk, Mamdouh. "Measuring sentences similarity: a survey." *arXiv preprint arXiv:1910.03940* (2019).

[10] Lee, Ming Che, Jia Wei Chang, and Tung Cheng Hsieh. "A grammar-based semantic similarity algorithm for natural language sentences." The Scientific World Journal 2014 (2014).

[11] Lee, Ming Che, et al. "Sentence similarity computation based on POS and semantic nets." 2009 Fifth International Joint Conference on INC, IMS and IDC. IEEE, 2009.

[12] Abdalgader, Khaled, and Andrew Skabar. "Short-text similarity measurement using word sense disambiguation and synonym expansion." *Australasian joint conference on artificial intelligence*. Springer, Berlin, Heidelberg, 2010.

[13] Schmidhuber, Jürgen. "Deep learning." Scholarpedia 10.11 (2015): 32832.

[14] Pontes, Elvys Linhares, et al. "Predicting the semantic textual similarity with siamese CNN and LSTM." *arXiv preprint arXiv:1810.10641* (2018).

[15] Li, Yulong, Dong Zhou, and Wenyu Zhao. "Combining Local and Global Features Into a Siamese Network for Sentence Similarity." *IEEE Access 8* (2020): 75437-75447.

[16] Medsker, Larry R., and L. C. Jain. "Recurrent neural networks." *Design and Applications* 5 (2001).

[17] Hochreiter, Sepp, and Jürgen Schmidhuber. "Long short-term memory." *Neural computation* 9.8 (1997): 1735-1780.

[18] Chung, Junyoung, et al. "Empirical evaluation of gated recurrent neural networks on sequence modeling." *arXiv preprint arXiv:1412.3555* (2014).

[19] Kenter, Tom, and Maarten De Rijke. "Short text similarity with word embeddings." *Proceedings of the 24th ACM international on conference on information and knowledge management*. 2015.

[20] Mikolov, Tomas, et al. "Distributed representations of words and phrases and their compositionality." *arXiv preprint arXiv:1310.4546* (2013).

[21] Mikolov, Tomas, et al. "Efficient estimation of word representations in vector space." *arXiv preprint arXiv:1301.3781* (2013).

[22] Heo, Tak-Sung, et al. "Sentence similarity evaluation using Sent2Vec and siamese neural network with parallel structure*." Journal of Intelligent & Fuzzy Systems Preprint:* 1-10.

[23] Ham, Jiyeon, et al. "KorNLI and KorSTS: New benchmark datasets for Korean natural language understanding." *arXiv preprint arXiv:2004.03289* (2020).